\def\year{2022}\relax
\documentclass[letterpaper]{article} % DO NOT CHANGE THIS
\usepackage{aaai22}  % DO NOT CHANGE THIS
\usepackage{times}  % DO NOT CHANGE THIS
\usepackage{helvet}  % DO NOT CHANGE THIS
\usepackage{courier}  % DO NOT CHANGE THIS
\usepackage[hyphens]{url}  % DO NOT CHANGE THIS
\usepackage{graphicx} % DO NOT CHANGE THIS
\def\UrlFont{\rm}  % DO NOT CHANGE THIS
\usepackage{natbib}  % DO NOT CHANGE THIS AND DO NOT ADD ANY OPTIONS TO IT
\usepackage{caption} % DO NOT CHANGE THIS AND DO NOT ADD ANY OPTIONS TO IT
\DeclareCaptionStyle{ruled}{labelfont=normalfont,labelsep=colon,strut=off} % DO NOT CHANGE THIS
\usepackage{algorithm}
\usepackage{algorithmic}

\usepackage{newfloat}
\usepackage{listings}
\floatstyle{ruled}
\newfloat{listing}{tb}{lst}{}
\floatname{listing}{Listing}
\usepackage{comment}
\usepackage{adjustbox}
\usepackage{multirow}

\title{Restore from Restored: Single-image Inpainting}

\author{
    Eunhye Lee\equalcontrib,
    Jeongmu Kim\equalcontrib,
    Jisu Kim\equalcontrib,
    Tae Hyun Kim\thanks{Tae Hyun Kim is corresponding author.}
}
\affiliations{
    Dept. of Computer Science, Hanyang University\\
    Seoul, South Korea\\
    $\{$dldms1345, jmkim1503, atat1270, taehyunkim$\}$@hanyang.ac.kr
}

\begin{document}

\maketitle

\begin{abstract}
Recent image inpainting methods have shown promising results due to the power of deep learning, which can explore external information available from the large training dataset.
However, many state-of-the-art inpainting networks are still limited in exploiting internal information available in the given input image at test time.
To mitigate this problem, we present a novel and efficient self-supervised fine-tuning algorithm that can adapt the parameters of fully pre-trained inpainting networks without using ground-truth target images.
We update the parameters of the pre-trained state-of-the-art inpainting networks by utilizing existing self-similar patches (i.e., self-exemplars) within the given input image without changing the network architecture and improve the inpainting quality by a large margin.
Qualitative and quantitative experimental results demonstrate the superiority of the proposed algorithm, and we achieve state-of-the-art inpainting results on publicly available benchmark datasets.
\end{abstract}

\section{Introduction}

\begin{figure}[t]
    \centering
    \includegraphics[width=1\columnwidth]{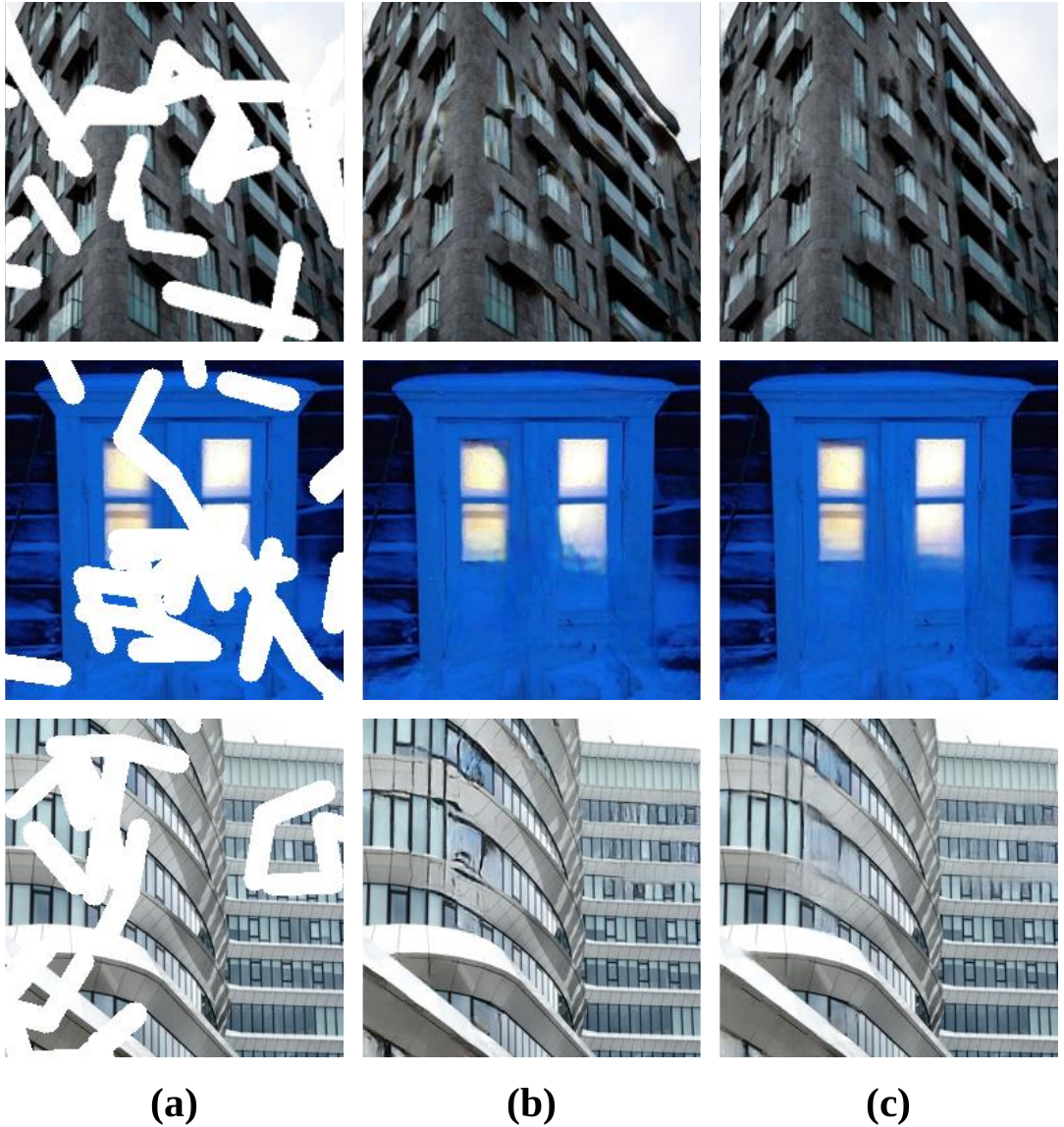}
    \caption{\small
    Our fine-tuning results with numerous inpainting networks.
    (a) Masked input images.
    (b) Initially restored results from pre-trained inpainting networks in the order of GatedConv~\cite{yu2019free}, EdgeConnect~\cite{nazeri2019edgeconnect}, and GMCNN~\cite{wang2018image}.
    (c) Our fine-tuning results.}
    \label{fig:intro}
\end{figure}

\label{sec:introduction}
Image inpainting, an image post-processing technique that removes unnecessary parts of images, such as subtitles and obstacles, and fills these areas with new colors and textures.
Inpainting has extensive applications in areas of photo editing, restoration, and even video editing~\cite{barnes2009patchmatch, levin2004seamless, yu2019free, kim2019deep}, and the main goal is to generate plausible images that are semantically consistent with the overall context and to fill the missing area to be continuous with the surrounding area.

Many traditional inpainting approaches based on the diffusion technique compute pixel values of the missing area using pixel values in the surrounding area~\cite{bertalmio2000image, ballester2001filling}.
Although diffusion-based approaches are effective in recovering small areas (e.g., subtitle removal), they produce blurry results when the missing area is large (e.g., object removal).
As an alternative, patch-match-based methods have been proposed to locate similar patches within the given image and copy intensity values of these similar patches into the missing area~\cite{liang2001real, hays2007scene, barnes2009patchmatch}.
These methods can fill the missing area with realistic contents using the information within the input image.
However, they cannot restore faces or complex landscapes that require information outside the input image.
%, and can create irrelevant structures due to the lack of image context inference capability.

The development of deep learning technologies has remarkably improved the performance of image inpainting techniques and made it possible to reconstruct considerably challenging images.
First, context encoders pioneered the use of deep neural networks in the image inpainting task and introduced an architecture composed of convolutional encoder and decoder with adversarial loss~\cite{pathak2016context, goodfellow2014generative}.
The encoder computes a latent feature representation from the input image and the decoder uses the feature representation to restore pixels in the missing area, consequently inferring and reconstructing the overall structure of the input image.
The output demonstrates a realistic texture as it approaches the distribution of the real image through the adversarial loss.
Follow-up studies carry out various attempts while maintaining encoder-decoder structures with discriminators, presented in the work of context encoders.
Among them, attention modules are exploited to generate realistic textures~\cite{yu2018generative, ren2019structureflow, li2020recurrent}.
These attention modules allow searching non-local regions to utilize information among similar structures in the given input image and fill pixel values of missing areas. 
Some studies perform structural restoration before carrying out the inpainting to clarify the boundary of the missing hole~\cite{ren2019structureflow, nazeri2019edgeconnect, xiong2019foreground}.

Although these neural approaches trained in a supervised manner with a large external training dataset for the inpainting task have shown promising and satisfying results, they have difficulties in fully utilizing the information within given test images.
Thus, we propose a new self-supervised learning approach to utilize internal statistics available within the given input test image such as patch-recurrence to overcome this limitation.
Patch-recurrence is a property that many similar patches are existing within a single natural image and it is utilized in numerous image restoration techniques including image denoising, inpainting, and super-resolution~\cite{michaeli2014blind, glasner2009super, huang2015single}.
In image inpainting, Deep Image Prior (DIP)~\cite{ulyanov2018deep} exploits self-similarity at the test time by training a randomly initialized neural network in a self-supervised manner.
However, its inpainting results highly depending on the selection of hyper-parameters, particularly for a large mask. % and take a long time in optimization.
Moreover, DIP is limited in generating plausible objects because it cannot exert the power of deep learning through large external datasets.
%We solve these problems by exploiting pre-trained inpainting networks. 

To alleviate these problems, we develop a new learning algorithm that combines supervised and self-supervised approaches to benefit from both the external dataset and the given input test image.
This approach improves the performance of existing state-of-the-art inpainting networks by simply updating network parameters using the internal information (i.e., repeating structure/texture and color distribution) available from the given input image at test time.
Specifically, we use parameters of fully pre-trained inpainting networks on a large external dataset as initial values.
We then fine-tune the network parameters in a self-supervised manner at the test stage by exploiting self-exemplars within the input test image and produce improved results as shown in Figure~\ref{fig:intro}.
Note that broken lines and crushed edges in the initial results are clearly improved, and the original shapes are properly restored with the proposed fine-tuning algorithm.
Moreover, our learning method is not restricted to specific network architecture and can be applied to various conventional networks and can easily upgrade the parameters of the conventional inpainting networks without changing their original architectures.
%A simple and effective test-time parameter update method that upgrades inpainting networks by combining self-supervision and supervision is proposed in this study.

The main contributions are summarized as follows:
\begin{itemize}
	\item A novel self-supervised fine-tuning method that exploits recurring patches within the test image is proposed.
%	\vspace{-6pt}
	\item Superior inpainting results on benchmark datasets are achieved by utilizing both internal and external datasets.
%	\vspace{-6pt}
	\item Our approach can be applied to various inpainting networks without modifying their original network architecture and loss functions.
\end{itemize}

\section{Related Work}
\label{sec:related}
%We review numerous methods for image inpainting as well as self-supervised learning approaches for image restoration relevant to the proposed approach.

\subsection{Image inpainting}
\begin{comment}
%The methods for image inpainting can be divided into the two approaches of traditional and learning-based methods.
Traditional image inpainting methods fill the missing area by progressively propagating information from the surrounding area or selecting highly similar patches within the given input image~\cite{bertalmio2000image,levin2003learning,Darabi12:ImageMelding12, hays2007scene}.
In particular, patch-based methods~\cite{Darabi12:ImageMelding12, hays2007scene,  huang2014image} can take advantage of not only local information but also non-local information and can fill the relatively large hole with meaningful content.
However, these methods struggle to create unique content, such as human faces or complex scenes, because they need to meet the assumption that several patches similar to the missing hole in the image are available.
\end{comment}
Recently, deep learning methods have shown successful results in image inpainting.
By learning the information of large external datasets, they overcome the limitation of traditional methods of trying to create unique content, such as human faces or complex scenes~\cite{bertalmio2000image,barnes2009patchmatch,huang2014image}.
Context encoder~\cite{pathak2016context} introduces an encoder-decoder network to extract the semantics of the image based on GAN to generate realistic details.
%Iizuka et al.~\cite{iizuka2017globally} removed the pooling layer and employed dilated convolutions~\cite{yu2015dilated} in the neural networks to solve the problem of the context encoder that generates blurry outputs because of the reduction of the resolution during the encoding stage.
%Also, they used two discriminators to maintain continuity in and out of the hole.
%The local one focus on the missing area and the global one assess the entire image.
Jiahui Yu~\shortcite{yu2018generative} propose the contextual attention module to exploit image contents far from the hole.
The module finds the most similar background patches with pixels of the missing area which are estimated in advance by a coarse network to reconstruct the missing area.
Generative multi-column CNN~\cite{wang2018image} uses parallel networks with different receptive field sizes to prevent the error propagation caused by the coarse and fine networks that are connected in series.
%It also uses the reconstruction loss that gives different weights to the pixels in the hole considering the distance from the boundary to solve the spatial-variant constraints.
Guilin Liu~\shortcite{liu2018image} proposed the use of partial convolution to learn various forms of masks, not only squared masks.
%Previous methods could not be utilized in practical cases where images were damaged in various forms because they were typically learned with squared masks.
The partial convolution layer performs the mask update and weight re-normalization of the masked area to propagate only valid information of the input.
In a similar approach by Jiahui Yu~\shortcite{yu2019free}, gated convolution automatically finds a proper mask for the input and considers the gating value calculated across a spatial location of every layer and channel.
EdgeConnect~\cite{nazeri2019edgeconnect} introduces a two-stage inpainting network model that first generates the edge map and then recovers the whole image.
However, this model is limited by its inability to utilize additional useful information, such as image color.
StructureFlow~\cite{ren2019structureflow} uses a smooth image while preserving the edges obtained via RTV~\cite{xu2012structure} instead of the edge map and employs appearance flow module~\cite{zhou2016view} to achieve an attention effect for realistic texture.
Jingyuan Li~\shortcite{li2020recurrent} presented recurrent feature reasoning (RFR) network that fills the missing area gradually to cover large holes using the knowledge consistent attention.

\begin{figure*}[t]
    \centering
    \includegraphics[width=1.0\linewidth]{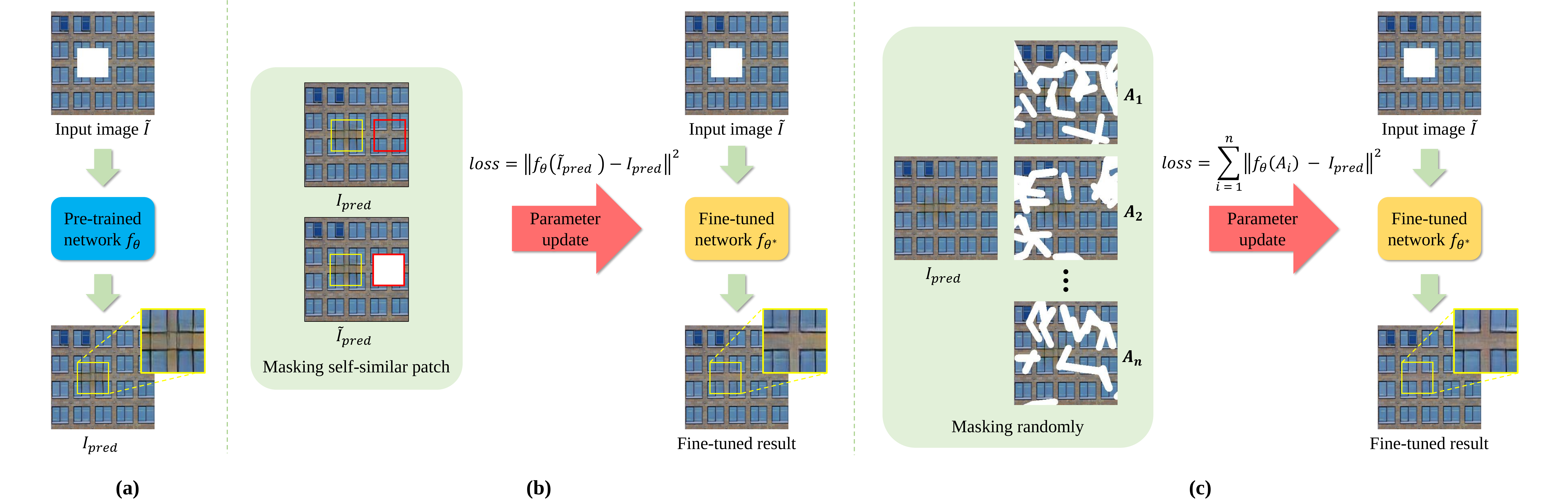}
    \caption{\small Inpainting results by EdgeConnect~\cite{nazeri2019edgeconnect} before and after fine-tuning.
    (a) Fully pre-trained EdgeConnect produces artifacts.
    (b) Fine-tuned EdgeConnect using a corresponding patch (red box) within the input can remove artifacts.
    (c) Fine-tuned EdgeConnect using randomly masked input can also remove artifacts without finding correspondences.
    }
    \label{fig:patchrecurrence}
\end{figure*}
\subsection{Self-supervised learning for restoration}
The self-supervised approach for image restoration primarily refers to the training of the neural networks using only information from the input images themselves without corresponding clean images~\cite{lehtinen2018noise2noise, batson2019noise2self, shocher2018zero}.
For the inpainting task, Deep Image Prior~\cite{ulyanov2018deep} trains a simple CNN using the internal statistics such as patch-recurrence and obtains the improved result without the ground-truth image.
However, these methods can only utilize internal data because they train the network from random scratch at the test time.

%We present that the self-supervised approach can be applied to image inpainting to fine-tune parameters of networks during test time by exploiting internal statics of a given masked input image.
In contrast, we not only exploit the internal information in the test input image but also utilize the information from the large external dataset similar to~\cite{park2020fast,lee2020restore} to overcome the limitation of the self-supervised approach that cannot exert the power of deep learning via large external datasets.

\section{Proposed Method}
In this section, we introduce a simple, yet effective self-supervised learning approach to adapt the parameters of pre-trained inpainting networks to the given input image.

\subsection{Patch-recurrence for inpainting}
The internal information available within the input image is very important for the image inpainting task and should be considered to generate semantically consistent and realistic results when filling missing areas.
Images generally demonstrate the property of patch recurrence, which enables the repeated presentation of many identical or similar patches in a single image.
The information of similar patches inside the image is used in past patch-match-based methods~\cite{Darabi12:ImageMelding12, barnes2009patchmatch, huang2014image} and recent deep-learning methods with the attention mechanism~\cite{yu2018generative,ren2019structureflow,li2020recurrent}.
Patch-match-based approaches find similar patches within the input to fill the hole and paste them.
By comparison, the attention module~\cite{yu2018generative, li2020recurrent} and appearance flow~\cite{ren2019structureflow} in deep approaches find similar patches/features in the given input image corresponding to the target (missing) area, and resulting attention maps are used to complete the texture within that area.
These methods improve the inpainting results by exploiting the internal information during test time by using additional network modules (e.g., non-local operator) to compute attention maps.
However, the search space of these modules for matching is relatively limited in practice, and cannot fully exploit the input image.
%and are inefficient because they require high computational costs to measure similarity among features at every pixel location. Specifically, $M \times N$ comparisons are required when the number of pixel values within missing and non-missing areas is $N$ and $M$, respectively, but only a few matches of them produce meaningful high-similarity scores.

Therefore, we introduce a new approach that can fully utilize the recurring patches in the given input and also present a new self-supervised fine-tuning method that enhances the quality of inpainting results of existing networks using these self-exemplars.

\subsection{Patch-match-based inpainting without explicit patch-match}

\begin{figure*}[t]
    \centering
    \includegraphics[width=\textwidth]{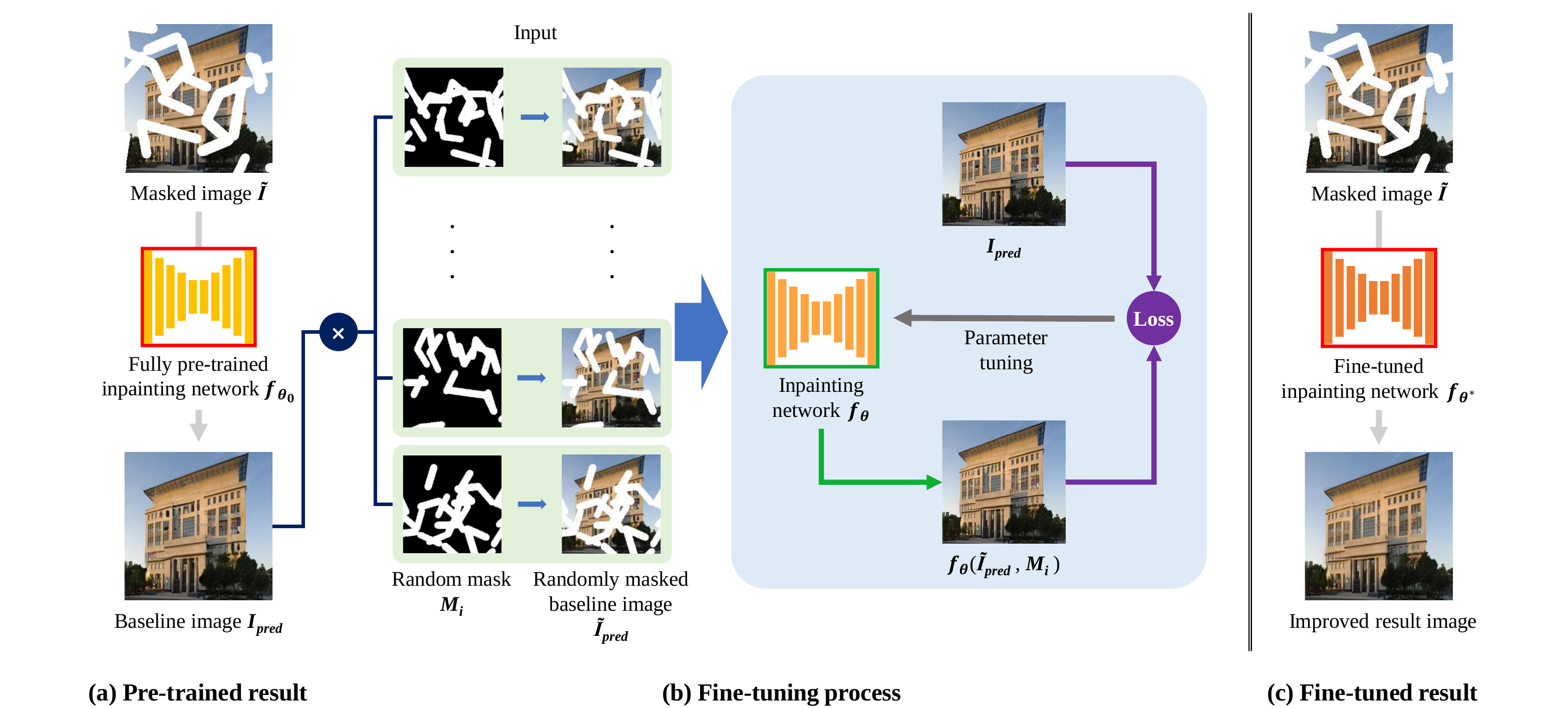}
    \caption{\small Overall flow of the proposed fine-tuning mechanism.
    (a) Initial inpainting result from the masked input image using the pre-trained network.
    (b) Fine-tuning with random masks on the result from the baseline network.
    (c) Fine-tuned network produces an improved result consistent with the input image.}
    \label{fig:process}
\end{figure*}

Existing learning-based inpainting methods learn how to reconstruct the missing area using a large number of training images during the training phase.
However, the input image given at the test phase may include unique and unseen information, such as unusual structures, textures, and colors, which are rare in the training dataset.
In this case, the pre-trained networks may generate unnatural outputs incompatible with the specific pattern in the background.
Therefore, we propose a new inpainting approach to exploit the specific information available within the input test image by utilizing multiple similar patches in the input and further improve the pre-trained networks by adapting the network parameters to the specific input image.

We illustrate how self-similar patches can be used to fine-tune the network parameters and improve the performance of conventional inpainting networks.
In Figure~\ref{fig:patchrecurrence}(a), we first obtain an initial inpainting result $I_{pred}$ by applying the fully pre-trained EdgeConnect~\cite{nazeri2019edgeconnect} network $f_{\theta}$ to an input image $\tilde{I}$ distorted with a squared mask.
Next, in Figure~\ref{fig:patchrecurrence}(b), we generate a newly masked image $\tilde{I}_{pred}$ to utilize repeating structures of the given input by removing a self-similar patch in $I_{pred}$ (red squared region).
Then, by minimizing the MSE between $I_{pred}$ and the restored image $f_{\theta}(\tilde{I}_{pred})$ from the newly masked image $\tilde{I}_{pred}$, we can fine-tune the network parameters $\theta$ to $\theta^*$.
Accordingly, we can update network parameters to the input image and can improve the performance of the inpainting network.
Note that we adapt network parameters without using the ground-truth image during the fine-tuning.
In Figure \ref{fig:patchrecurrence}(c), we show that we can also fine-tune the network by corrupting the initially restored image with random masks without explicit patch-match to find self-similar patches.
Specifically, we generate multiple training images for fine-tuning by corrupting $I_{pred}$ using random masks, then fine-tune the network by minimizing MSE between the initial inpainting result $I_{pred}$ and the restored image $f_{\theta}(A_i)$ from the newly corrupted image $A_i$.
By doing so, we demonstrate that we can adapt network parameters without explicitly searching for self-similar patches within the input as in Figure \ref{fig:patchrecurrence}(b).

Conventional methods require dedicated algorithms and specialized network modules to search similar patches explicitly~\cite{barnes2009patchmatch, Darabi12:ImageMelding12, yu2018generative, ren2019structureflow}.
In contrast, we learn the self-exemplars by using the randomized masking (corruption) scheme during the test phase without any additional algorithm or network modules to find similar patches (e.g., patch-match).
As the area of similar patches can be randomly masked several times during the randomized masking scheme, these self-similar patches can be naturally exposed to the network during the fine-tuning stage.
Thus, the inpainting network can learn to restore the missing area using similar patches, as shown by the result of Figure~\ref{fig:patchrecurrence}(c), which shows similar improvements to Figure~\ref{fig:patchrecurrence}(b).
%Another advantage of this approach is that, unlike conventional approaches, we don't need to restrict the shape, size, and number of the recurring patches for the randomized-masking-based fine-tuning mechanism.
Notably, another advantage of this approach is that the randomized-masking-based fine-tuning mechanism does not restrict the shape, size, and number of the recurring patches in the input.

\begin{algorithm}[t]
	\textbf{Input:} input masked image $\tilde{I}$, original mask $M$
	
	\textbf{Require:} inpainting network $f$ and the pre-trained parameter $\theta_0$,   number of training $T$, random masks $\{M_{i}\}$,
	learning rate $\alpha$
	
	\textbf{Output:} enhanced inpainting result $f_{\theta^*}(\tilde{I}, M)$
	
	\begin{algorithmic}[1]

	\STATE $i$ $\leftarrow$ 0
	
	\STATE $\theta \leftarrow \theta_0$
	
	\STATE $I_{pred} \leftarrow f_{\theta_0}(\tilde{I}, M)$
	
	\WHILE{i $<$ T}
	    
	    \STATE $\tilde{I}_{pred} \leftarrow I_{pred}\odot \left ( 1 - M_i \right )$
	    
	    \STATE $I_{pred\left (\theta \right )} \leftarrow f_{\theta}(\tilde{I}_{pred}, M_{i})$
	    %\nl $I_{restored} \leftarrow f_{\theta}(\tilde{I}_{base}, M_{i})$
	    
	    \texttt{\\}

		%\nl $loss_|{1}(\theta) \leftarrow loss_{em}(I_{pred}\odot(1-M)$,
		%\\ \qquad \qquad \qquad \qquad
		%$I_{pred(\theta)}\odot(1-M)|^2$
		\small // Random transformations can be applied for the loss
		\STATE $loss_{rec}(\theta) \leftarrow \|I_{pred}\odot(1-M)-I_{pred(\theta)}\odot(1-M)\|^2$
		
		\texttt{\\}
		
		%\nl $loss_{adv}(\theta) \leftarrow loss_{other}(I_{pred}, I_{pred(\theta)})$
		\STATE $loss_{adv}(\theta) \leftarrow {\small VGG~and/or~ adversarial~losses}$
		
		%\nl
		\STATE $Loss(\theta) \leftarrow loss_{rec}(\theta) + loss_{adv}(\theta)$
		
        \texttt{\\}
		
		\small // Parameter update
		\STATE $\theta \leftarrow \theta - \alpha \nabla_{\theta}  \textit{Loss}(\theta)$

        %\nl
		\STATE $i$ $\leftarrow $ $i$ + 1  
	\ENDWHILE

	\STATE $\theta^* \leftarrow \theta$

	\STATE \textbf{Return:}  $f_{\theta^*}(\tilde{I}, M)$
	\caption{\small \textit{Fine-tuning algorithm}}
	\label{algorithm}
	\end{algorithmic}
\end{algorithm}
\subsection{Overall flow}\label{overall_flow}

The overall flow of the proposed fine-tuning approach is described in Algorithm~\ref{algorithm} and illustrated in Figure~\ref{fig:process}.

We first start the fine-tuning process with the initially restored image $I_{pred}$ by using the fully pre-trained inpainting network as follows:
\begin{equation}
    I_{pred} = f_{\theta_0} (\tilde{I}, M ),
    \label{eq3}
\end{equation}
where $\theta_0$ denotes the fully pre-trained parameters of the baseline inpainting network $f$ and $\tilde{I}$ denotes the masked input image.
A binary map $M$ denotes a given input mask where missing and other areas are represented by 1 and 0.

Second, we acquire a training dataset using the initially restored image $I_{pred}$.
We generate a randomly corrupted masked image for the $i_{th}$ fine-tuning iteration as follows:
\begin{equation}
    \tilde{I}_{pred} = I_{pred}\odot \left ( 1 - M_i \right ),
    \label{eq5}
\end{equation}
where $M_i$ denotes the randomly generated binary mask and $\odot$ represents the element-wise multiplication.
The newly synthesized masked image $\tilde{I}_{pred}$ and initially restored image $I_{pred}$ become the input and target of our training dataset.
We render a restored image $I_{pred\left (\theta \right )}$ from the newly masked image $\tilde{I}_{pred}$ using the inpainting network $f_{\theta}$.

Third, we compute gradient values with respect to network parameters using the predefined loss functions used in pre-training the baseline inpainting network and then update network parameters using a conventional optimizer (e.g., ADAM), and the loss is computed using $I_{pred}$ and $I_{pred\left (\theta \right )}$.
Notably, if the baseline inpainting network includes GAN architecture, the discriminator of GAN computes the adversarial loss by using $I_{pred}$ as the real sample since the ground-truth clean image is not available.

%We improve the performance by making slight modifications in calculating the loss. The generated part of $I_{pred}$ can affect performance, so we exclude that part from the image when calculating the loss, if possible.
%When calculating the pixel-wise losses (i.e., reconstruction loss), such as L2, we exclude the part of the original mask $M$, and the expression is as follows:
For instance, for the pixel-wise loss (i.e., reconstruction loss), we can use the L2 loss function as follows:
\begin{equation}
%    loss_{1}(\theta) = loss_{em}(I_{pred}\odot(1-M), I_{pred(\theta)}\odot(1-M)).
    loss_{rec}(\theta) = \|I_{pred}\odot(1-M)- I_{pred(\theta)}\odot(1-M)\|^2.
    \label{ExcludeMask}
\end{equation}
In practice, to compute the reconstruction loss, we exclude the part corresponding to the original mask $M$ (i.e., initially restored area) since we see improvements in our experiments by making this slight modification. Note that we can use any conventional reconstruction losses (e.g., L1).
Moreover, when considering additional losses, such as VGG and adversarial losses, we do not need to make any modifications. We employ the original loss functions used to pre-train the baseline inpainting networks, 
and the expression is $loss_{adv}(\theta)$.
%as follows:
%\begin{equation}
%    loss_{2}(\theta) = loss_{other}(I_{pred}, I_{pred(\theta)}).
%    \label{OtherLoss}
%\end{equation}
Thus, the overall loss function that updates parameters of the given inpainting network is expressed as follows:
\begin{equation}
    Loss\left ( \theta \right ) = loss_{rec}(\theta) + loss_{adv}(\theta).
    \label{OverallLoss}
\end{equation}

Then, we repeat these steps $T$ times.
%, which is the number of iterations determined experimentally to achieve best results. 
Notably, random transformations can be applied during fine-tuning to prevent the network from producing $I_{pred}$ regardless of its input.

Finally, the network parameter is upgraded to $\theta^*$, and we obtain the final fine-tuned image ${f}_{\theta^*}(\tilde{I}, M)$ with the original input image $\tilde{I}$ and mask $M$.

%We call this algorithm ``restore-from-restored" because this method leverages the already restored image to improve the inpainting performance.

\section{Experimental result}
Please refer to our supplementary material for more results, and the code will be publicly available upon acceptance.
\subsection{Implementation details}
To evaluate the performance of the proposed fine-tuning algorithm, we use three different inpainting networks,  GatedConv~\cite{yu2019free}, EdgeConnect~\cite{nazeri2019edgeconnect}, and GMCNN~\cite{wang2018image}, as baseline networks of our algorithm.
The experiments are performed using the official code for each model.
%Notably, EdgeConnect is currently a state-of-the-art inpainting network.

For our experiments, we use officially available and fully pre-trained network parameters on the Places2 dataset~\cite{zhou2017places} for each network and fine-tune the pre-trained parameters via the proposed method in Algorithm.~\ref{algorithm}.
% EdgeConnect provide the generator and discriminator parameters, GMCNN and GatedConv only provide the generator parameters, and we used only given values without learning from scratch.
We evaluate the performance of the proposed algorithm on test sets in conventional benchmark datasets, such as Places2 and Urban100~\cite{huang2015single}.
The Places2 dataset is a mixed dataset of people, landscapes, and buildings; the Urban100 dataset mainly consists of buildings with repetitive patterns.
%We generate training inputs to fine-tune the models at the test time by applying random free-form masks~\cite{yu2019free} to the initially restored image $I_{pred}$.
%We use the random free-form masks used in the fine-tuning process are obtained by the algorithm for sampling free-form masks introduced in \cite{yu2019free}.
We use the random free-form masks for fine-tuning introduced in \cite{yu2019free}.
The input image size for the fine-tuning is 256$\times$256 and each input image includes holes covering approximately 20\% to 40\% of the image.
%In the fine-tuning step, we use the same mask data that the authors used to pre-train each model.
We utilize the same optimizer and loss function used in each of the pre-trained models, but we modify the learning rate for the fine-tuning: we use learning rate of $10^{-5}$ for EdgeConnect and GatedConv, and $10^{-4}$ for GMCNN.

Our experiments are conducted with Intel i9 and NVIDIA RTX2080 Ti GPU, and it takes 0.2, 0.3, and 0.7 seconds to perform 1 fine-tuning iteration (patch-size = 256$\times$256, batch-size = 4) with GMCNN, GatedConv, and EdgeConnect respectively.

\subsection{Ablation study}\label{sec_ablation}
\paragraph{Number of fine-tuning iterations $T$ and image quality}

\begin{figure}[t]
    \centering
    \includegraphics[width=1.0\linewidth]{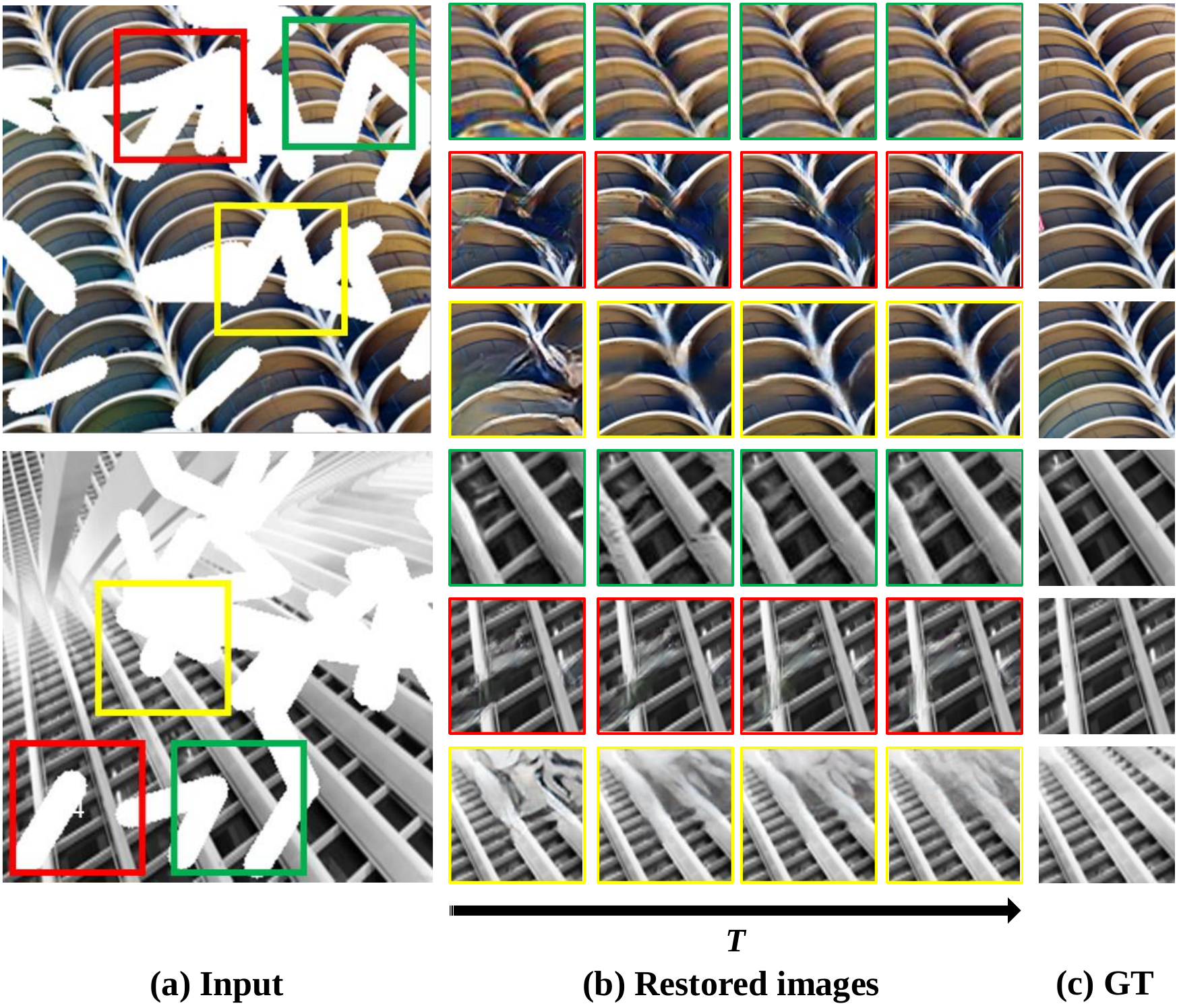}
    \caption{\small Visual results on the Urban100 dataset according to the fine-tuning progress. Green, red, and yellow boxes represent the results by GatedConv, EdgeConnect, and GMCNN, respectively.
    (a) Input masked images.
    (b) Initially restored image and fine-tuned images for $T$ iterations. 
    \textbf{Green box}: GatedConv results (From left to right: $T$=0, $T$=100, $T$=200, $T$=400). \textbf{Red box}: EdgeConnect results (From left to right: $T$=0, $T$=100, $T$=500, $T$=1000).
    \textbf{Yellow box}: GMCNN results (From left to right: $T$=0, $T$=50, $T$=150, $T$=200).
    (c) Ground-truth images.}
    \label{fig:iterprogress}
\end{figure}

\begin{figure*}[t]
    \centering
    \begin{adjustbox}{max width=0.95\linewidth}
    \includegraphics{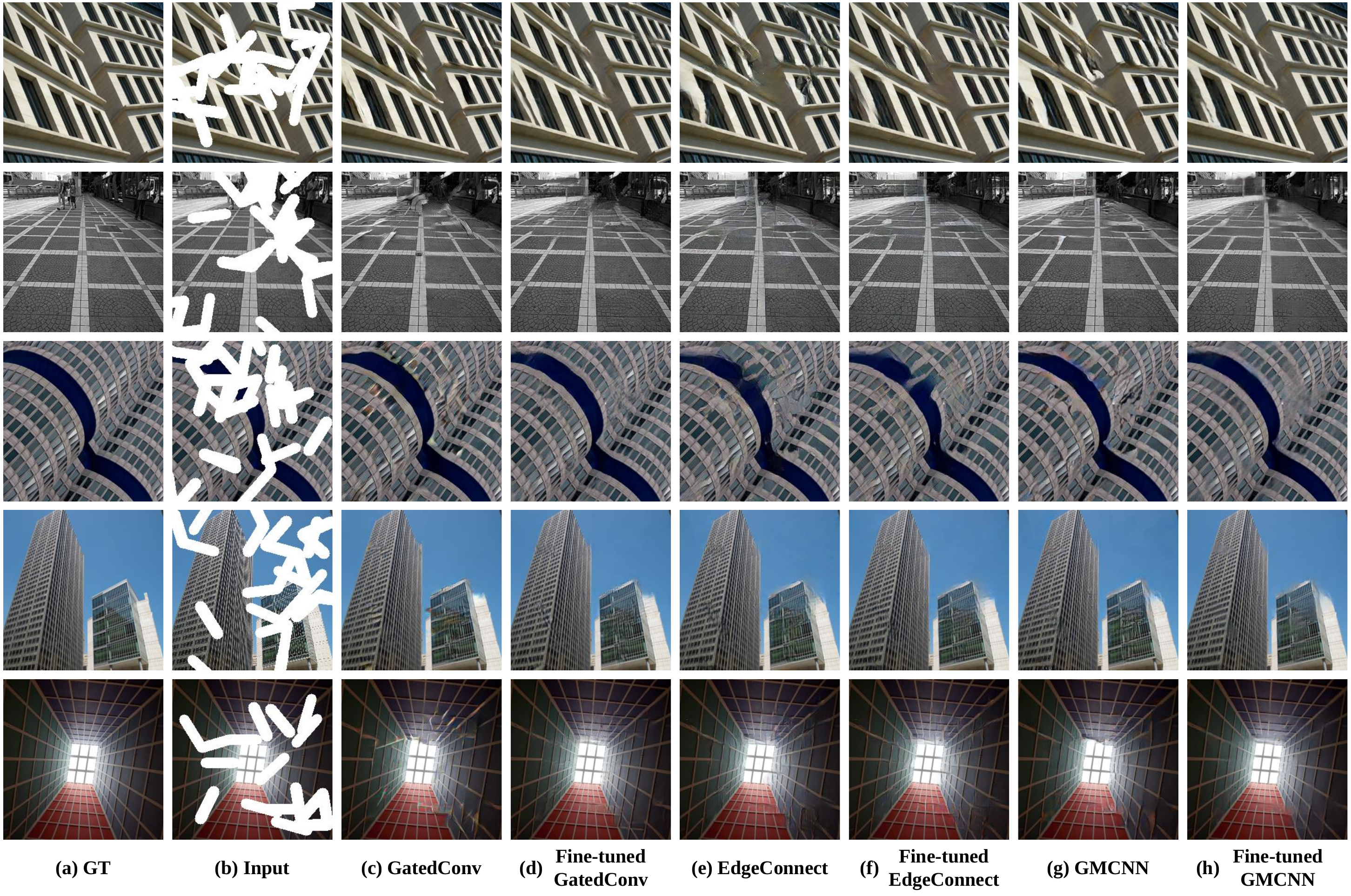}
    \end{adjustbox}
    \caption{\small 
    %Qualitative comparison of results on the Urban100 dataset before and after fine-tuning pre-trained models.
    %The first two columns show the ground-truth and masked input image.
    %The next two columns represent image pairs, where the left is the baseline image obtained via the pre-trained inpainting models and the right is the result images by our fine-tuning algorithm.
    Qualitative comparison of results on the Urban100 dataset.
    From left to right: ground-truth, the masked input image, three image pairs of before and after fine-tuning each model.
    }
    \label{fig:qualitative_urban}
    %\vspace{-2Ex}
\end{figure*}

In Figure \ref{fig:iterprogress}, we show changes in image quality depending on the number of fine-tuning iterations.
%Figure \ref{fig:iterprogress} illustrates the results of the fine-tuned network with different numbers of iterations for each pre-trained model.
The structure of the resulting image is gradually improved as fine-tuning progress.
%, and the result becomes close to the ground truth when it approaches to the pre-fixed iteration.
%However, fine-tuning beyond the appropriate number of iterations causes the model to overfit and produce artifacts or blurry images. Artifacts can occur in the early stage when networks can not fully utilize the self-similar patches.
% Table~\ref{tab:places2-table} shows.

However, we sometimes observe unexpected artifacts when $T$ is very small since networks can not fully utilize the self-similar patches, and see over-fitted and blurry images when $T$ is very large. 
Thus, we need to determine the optimal fine-tuning iterations to produce the best results, and we use the Fréchet inception distance (FID)~\cite{heusel2017gans} which can measure the image quality without the ground-truth clean image.
%To produce the best results through fine-tuning, we perform the small experiment for each model to choose optimal fine-tuning iterations.
Specifically, we compute FID score at each iteration and stop the fine-tuning when it starts to increase.
%We use random 30 test images in Places2 dataset to fine-tune models and measure the Fréchet inception distance (FID)~\cite{heusel2017gans} score of the fine-tuned results at each iteration number.
%We choose the best iteration for each model as the number that shows the lowest FID value. 
We find that FID values are highly consistent with human judgment for the quality of images produced by the same image.
Please refer to the supplementary for more details.
% fid로 재는 이유는 낮은 fid가 높은 psnr보다 사람이 고르는 best 이미지와 대체로 일치하기 때문이다. 실제 실험에서 확인해본 결과 fig(b)의 예제에서 보는 바와 같이 대부분의 이미지가 최적 iter에서 결과가 괜찮았다.
%We observe that a large number of iterations for the fine-tuning can lead to over-smoothed results depending on the model, and we demonstrate that the FID score can be an important criterion for determining the proper number of iterations $T$ because of its higher sensitivity to over-smoothing than the PSNR value.
% 모델마다 최적 iter가 다른 이유
In the remaining experiments, we use the fixed number of iterations based on this manner, and around 400, 1000, and 200 iterations for each pre-trained model (i.e., GatedConv, EdgeConnect, and GMCNN) are best on average.

% \begin{figure}[]
%     \centering
%     \includegraphics[width=1\linewidth]{figs/urbanG_final_cropped.pdf}
%     \caption{\small Improvement of FID and PSNR values of GatedConv, EdgeConnect, and GMCNN with the fine-tuning process on the Urban100 dataset.
%     High PSNR values and low FID scores indicate better quality of images.}
%     \label{fig:urbangraph}
% \end{figure}

\begin{table}[t]
\begin{adjustbox}{max width=1.0\linewidth}
\centering
\begin{tabular}{ccccccc}
\hline
\textbf{Model} & \textbf{\# iter.} (=$T$) & \textbf{PSNR} & \textbf{SSIM} & \textbf{FID} & \textbf{LPIPS} \\ \hline
\multirow{5}{*}{\textbf{\begin{tabular}[c]{@{}c@{}}Gated\\ Conv\end{tabular}}} & 0 (baseline) & 23.25 & 0.8831 & 73.93 & 0.103 \\
 & 100 & 23.10 & 0.8767 & 77.22 & 0.109 \\
 & 200 & 23.36 & 0.8820 & 73.55 & 0.103 \\
 & 400 & 23.76 & 0.8890 & \textbf{72.82} & \textbf{0.097} \\
 & 1000 & \textbf{23.97} & \textbf{0.8913} & 79.08 & 0.101 \\ \hline
\multirow{5}{*}{\textbf{\begin{tabular}[c]{@{}c@{}}Edge\\ Connect\end{tabular}}} & 0 (baseline) & 24.00 & 0.8934 & 28.06 & 0.098 \\
 & 100 & 24.04 & 0.8941 & 27.94 & 0.098 \\
 & 200 & 24.06 & 0.8946 & 27.78 & 0.097 \\
 & 400 & 24.09 & 0.8950 & 27.57 & 0.097 \\
 & 1000 & \textbf{24.12} & \textbf{0.8957} & \textbf{27.23} & \textbf{0.096} \\ \hline
\multirow{5}{*}{\textbf{GMCNN}} & 0 (baseline)& 23.79 & 0.8477 & 69.92 & 0.078 \\
 & 100 & \textbf{24.70} & \textbf{0.8586} & 65.56 & 0.077 \\
 & 200 & 24.61 & 0.8550 & \textbf{61.60} & \textbf{0.069} \\
 & 400 & 24.47 & 0.8488 & 67.44 & 0.070 \\
 & 1000 & 24.18 & 0.8439 & 75.13 & 0.082 \\ \hline
\end{tabular}
\end{adjustbox}
\caption{\small Fine-tuning results of various inpainting models on the Places2 dataset by changing the number of iterations.}
\label{tab:places2-table}
%\vspace{-3Ex}
\end{table}

\begin{table}[t]
\begin{adjustbox}{max width=1.0\linewidth}
\centering
\begin{tabular}{cccc}
\hline
\textbf{} & \begin{tabular}[c]{@{}c@{}}\textbf{GatedConv}\end{tabular} & \begin{tabular}[c]{@{}c@{}}\textbf{EdgeConnect}\end{tabular} & \begin{tabular}[c]{@{}c@{}}\textbf{GMCNN}\end{tabular} \\ \hline
\textbf{PSNR$^\star$} & 21.38 $\rightarrow$ 22.64 & 22.73 $\rightarrow$ 23.19 & 21.64 $\rightarrow$ 23.38 \\ \hline
\textbf{SSIM$^\star$} & 0.862 $\rightarrow$ 0.888 & 0.880 $\rightarrow$ 0.892 & 0.830 $\rightarrow$ 0.851 \\ \hline
\textbf{FID$^\dagger$} & 75.34 $\rightarrow$ 58.80 & 53.98 $\rightarrow$ 45.34 & 77.96 $\rightarrow$ 70.83 \\ \hline
\textbf{LPIPS$^\dagger$} & 0.096 $\rightarrow$ 0.078 & 0.085 $\rightarrow$ 0.073 & 0.083 $\rightarrow$ 0.082 \\ \hline
\end{tabular}
\end{adjustbox}
\caption{\small Comparisons between results before and after fine-tuning of each baseline model on Urban100 dataset.
Lower $^\dagger$ and higher $^\star$ scores indicate better quality of images.}
\label{tab:urban-table}
%\vspace{-2Ex}
\end{table}

\subsection{Quantitative results}

\begin{figure}[h]
    \centering
    \begin{adjustbox}{max width=1.0\linewidth}
    \includegraphics{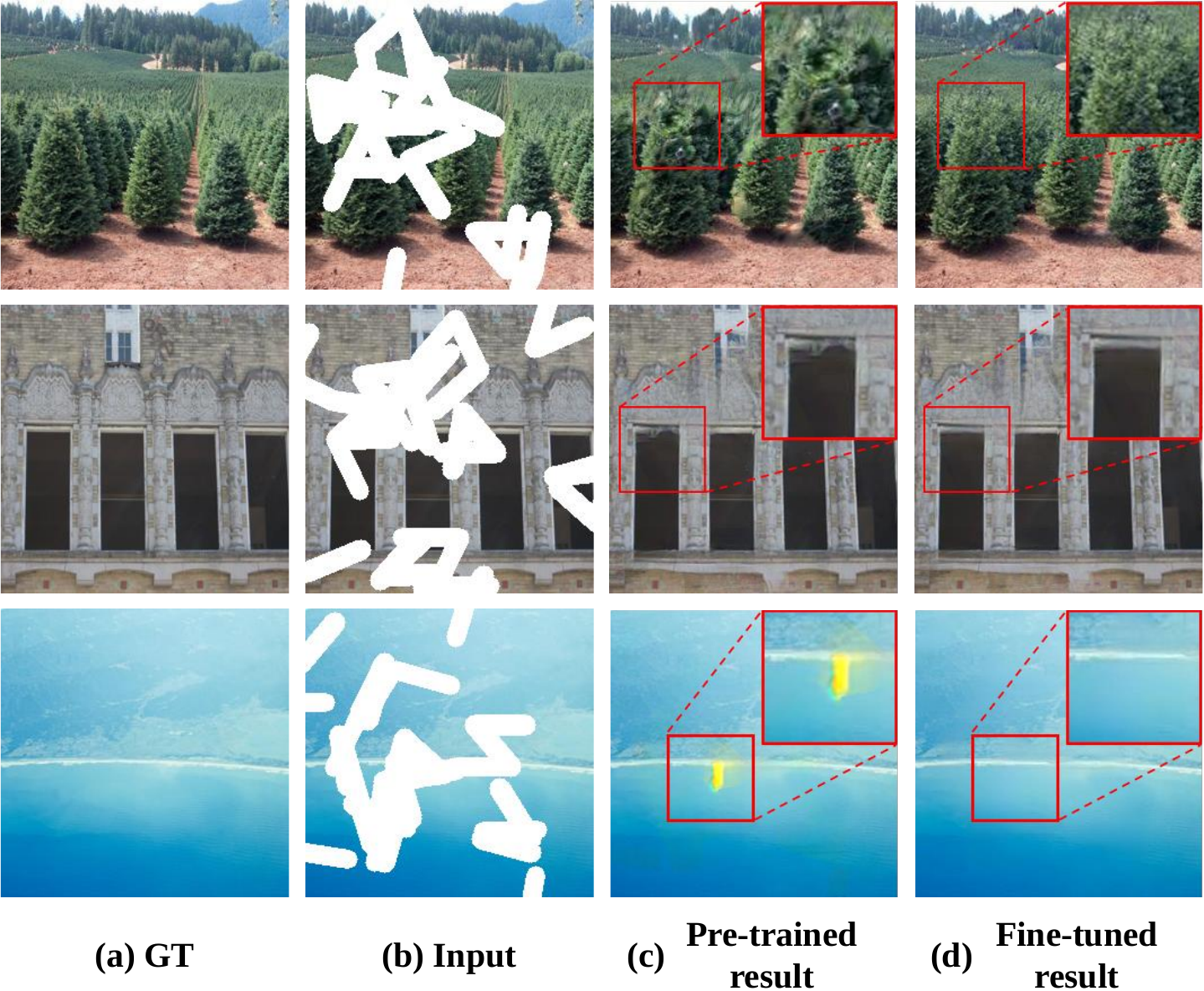}
    \end{adjustbox}
    \caption{\small Qualitative comparison of results on the Places2 dataset before and after fine-tuning the pre-trained models.
    }
    \label{fig:qualitative_places}
    %\vspace{-2Ex}
\end{figure}

We quantitatively evaluate the performance of fine-tuned networks with peak signal-to-noise ratio (PSNR), and structural similarity index measure (SSIM) to compare the inpainting results objectively.
These methods mainly measure the distortion of the results, assuming that the ideal results are the same as the original.
Moreover, we compare the perceptual quality of the results by comparing FID and LPIPS~\cite{zhang2018unreasonable},
%how plausible the produced image looks from the human perspective. This comparison is necessary because the inpainting task does not simply recover images quantitatively but furthermore produces visually appealing images to humans based on the given input image. We use learned perceptual image patch similarity (LPIPS)~\cite{zhang2018unreasonable} and Fréchet inception distance (FID)~\cite{heusel2017gans} values.% for the perceptual comparison.
%To be specific, LPIPS compares the deep features of images extracted from the image classifier based on a CNN architecture, such as VGG~\cite{simonyan2014vgg} and AlexNet~\cite{krizhevsky2014alex}.
and we use AlexNet~\cite{krizhevsky2014alex} to compute LPIPS in this study.
%FID measures the Wasserstein distance (Fréchet distance) between distributions of features from the real and generated images.
%The mean and covariance of each distribution are calculated through the fully pre-trained Inception-V3~\cite{szegedy2016rethinking} network.

Table~\ref{tab:places2-table} and Table~\ref{tab:urban-table} list the quantitative restoration results.
First, Table~\ref{tab:places2-table} shows the comparison of the results from each fine-tuning iteration on the Places2 dataset.
One thousand images from the Places2 test dataset are used for the evaluation.
With the aid of our fine-tuning algorithm, overall metrics values improve consistently compared with the restoration results by the pre-trained baseline models (i.e., $T$=0).
Next, Table \ref{tab:urban-table} presents the improvements between the results of pre-trained models and the outcomes after fine-tuning on the Urban100 test dataset.
Although the initial results from the baseline models are poor since the Urban100 dataset is not used in pre-training the baseline networks, our fine-tuning results show considerable improvements by test-time adaptation.
This finding proves that our fine-tuning method enhances the results although the input test image has a slightly different distribution from the dataset used in pre-training.

\subsection{Qualitative results}

We compare the qualitative results between the initially restored results by the pre-trained models and our results by fine-tuning the pre-trained baselines.
Figure~\ref{fig:qualitative_places} shows visual results on the Places2 test dataset.
Note that the generated part in the result images of pre-trained models is distorted or does not match the other part.
By comparison, our fine-tuned models generate more natural results.
Furthermore, the visual results are significantly improved when multiple repetitive patterns, such as windows and stairs, exist in the input image since the network is likely to learn the correct answer using many similar patches.
% Figure~\ref{fig:qualitative_urban} shows the fine-tuning results on the Urban100 dataset.
The results on the Urban100 test dataset demonstrate improved performance for structural restoration due to the property of containing many repetitive structures within the image as shown in Figure~\ref{fig:qualitative_urban}.
%More results are provided in the supplementary material.

\subsection{Results from the non-repetitive images}

\begin{figure}[t]
    \centering
    \includegraphics[width=0.95\columnwidth]{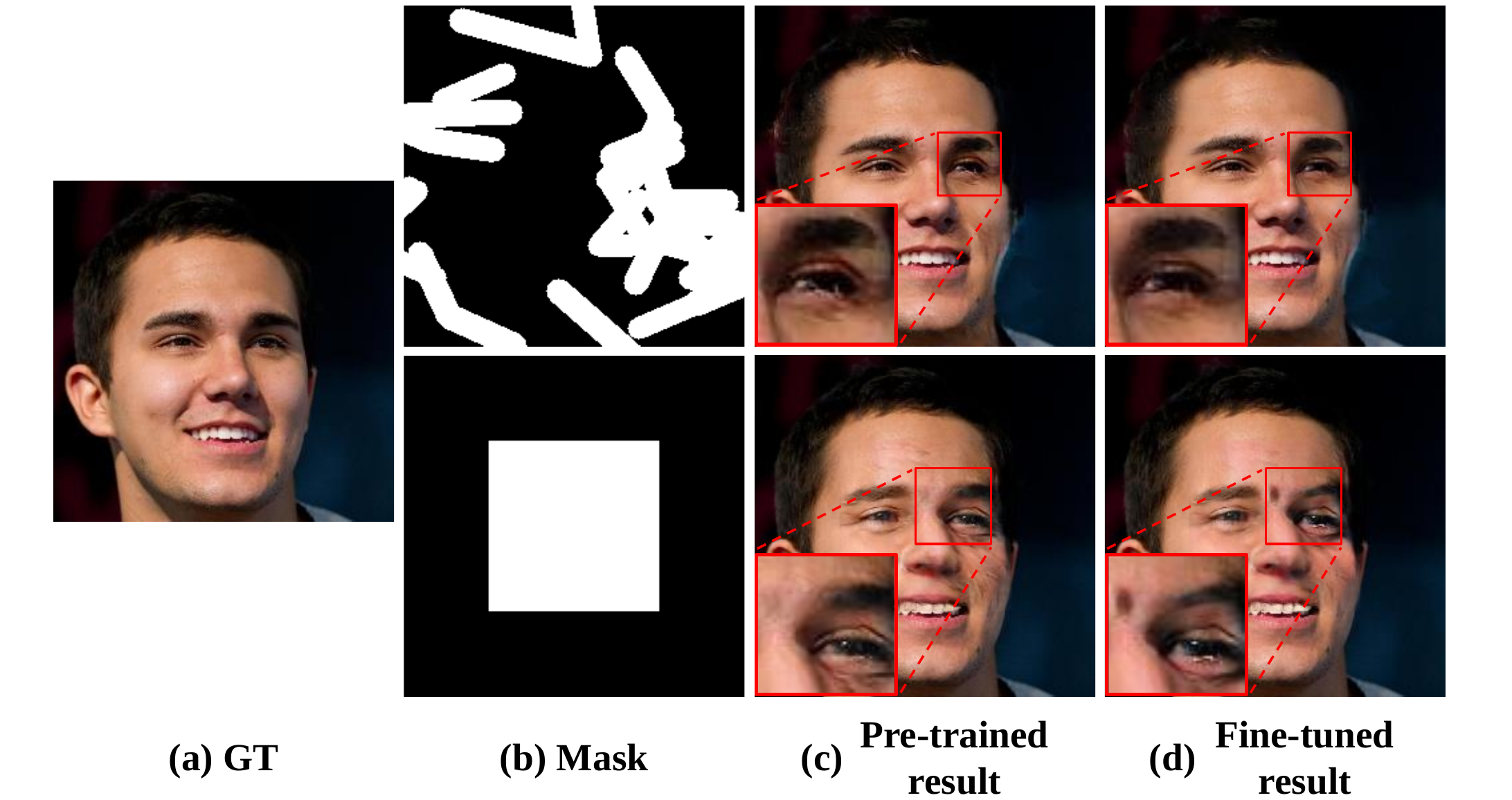}
    \caption{\small Results of Non-repetitive dataset.
    The result of the free-form mask that masks only one eye shows the improvement in the shape and color consistency of masked eye. On the other hand, the results of center mask shows the degradation in quality.}
    \label{fig:celeba_rs}
\end{figure}

% figure; best case, worst case of celeba-hq, gatedconv
We conduct experiments on Celeba-HQ~\cite{karras2017progressive} to show the result of our method from the non-repetitive facial input image with two different input masks: a central rectangle mask and a random free-form mask.
Celeba-HQ is a large dataset including human faces, and we choose random 500 test images.
We use the fully pre-trained GatedConv on the Celeba-HQ dataset.
Figure~\ref{fig:celeba_rs} shows the comparison of before and after fine-tuning with two different shapes of masks.
%when the input is corrupted by a free-form mask including one eye. 
We can see improved eye shape and skin color consistency after fine-tuning when the input is corrupted by a free-form mask that includes one eye because the fine-tuned model can learn the self-similarity of the remaining part. However, when the input image is corrupted by a rectangle mask that covers two eyes, we can see the degraded result after fine-tuning.
%the results using centeral masks that remove every unique part of the face(e.g. two eyes, nose, and mouth) show the lower quality after fine-tuning.
Table~\ref{tab:celebatest} indicates corresponding quantitative results.
% color or sth은 smoothing되어서 결과가 잘 나왔다, free-form 마스크 사용해서 참조할 부분이 남아있는 경우 결과가 괜찮았다.
% center mask로 눈코입 다 날려버린 경우 non-repetitive 요소들은 복구가 잘 되지 않았다.

\section{Conclusion}
A new self-supervision-based inpainting algorithm that allows the adaptation of fully pre-trained network parameters during the test stage is proposed.
We utilize self-similar patches within the given input test image to fine-tune the network without using the ground-truth clean image and elevate the performance of networks by combining internal and large external information.
We can easily fine-tune the baseline networks and significantly improve the performance over the baselines by optimizing loss functions, which are used to pre-train the baseline networks.
The proposed method achieves state-of-the-art inpainting results on the conventional benchmark datasets, and extensive experimental results demonstrate the superiority of our method.

\begin{table}[t]
\centering
\begin{tabular}{ccc}
\hline
 & \textbf{Free-form mask} & \textbf{Center mask} \\ \hline
\textbf{PSNR} & 26.53 $\rightarrow$ 27.04 & 25.64 $\rightarrow$ 25.55 \\ \hline
\textbf{FID} & 29.48 $\rightarrow$ 28.19 & 31.03 $\rightarrow$ 31.79 \\ \hline
\end{tabular}
\caption{Comparisons between results before and after fine-tuning by GatedConv model on Celeba-HQ dataset.}
\label{tab:celebatest}
\end{table}

\bigskip
\bibliography{aaai22}

\end{document}